\documentclass[10pt,twocolumn,letterpaper]{article}

\usepackage{cvpr}
\usepackage{times}
\usepackage{epsfig}
\usepackage{graphicx}
\usepackage{amsmath}
\usepackage{amssymb}
\usepackage{subfigure}
\usepackage{multirow}
\usepackage[table,xcdraw]{xcolor}
\usepackage{epstopdf}

\cvprfinalcopy 

\def\cvprPaperID{195} 

\newtheorem{theorem}{Theorem}

\ifcvprfinal\fi
\begin{document}


\title{ISTA-Net: Interpretable Optimization-Inspired Deep Network for Image Compressive Sensing}


\author{Jian Zhang, Bernard Ghanem\\
King Abdullah University of Science and Technology (KAUST), Saudi Arabia\\
{\tt\small jian.zhang@kaust.edu.sa, bernard.ghanem@kaust.edu.sa}
}

\maketitle
\thispagestyle{empty}

\begin{abstract}
\noindent With the aim of developing a fast yet accurate algorithm for compressive sensing (CS) reconstruction of natural images, we combine in this paper the merits of two existing categories of CS methods: the structure insights of traditional optimization-based methods and the speed of recent network-based ones. Specifically, we propose a novel structured deep network, dubbed ISTA-Net,  which is inspired by the Iterative Shrinkage-Thresholding Algorithm (ISTA) for optimizing a general $\ell_1$ norm CS reconstruction model. To cast ISTA into deep network form, we develop an effective strategy to solve the proximal mapping associated with the sparsity-inducing regularizer using nonlinear transforms. All the parameters in ISTA-Net (\eg nonlinear transforms, shrinkage thresholds, step sizes, etc.) are  learned end-to-end, rather than being hand-crafted. Moreover, considering that the residuals of natural images are more compressible, an enhanced version of ISTA-Net in the residual domain, dubbed {ISTA-Net}$^+$, is derived to further improve CS reconstruction. Extensive CS experiments demonstrate that the proposed ISTA-Nets outperform existing state-of-the-art optimization-based and network-based CS methods by large margins, while maintaining fast computational speed. Our source codes are available: \textsl{http://jianzhang.tech/projects/ISTA-Net}.
\vspace{-12pt}
\end{abstract}

\section{Introduction}
From much fewer acquired measurements than determined by Nyquist sampling theory, Compressive Sensing (CS) theory demonstrates that a signal can be reconstructed with high probability when it exhibits sparsity in some transform domain \cite{candes2006near, duarte2008single}. This novel acquisition strategy is much more hardware-friendly and it enables image or video capturing with a sub-Nyquist sampling rate \cite{sankaranarayanan2012cs, liutkus2014imaging}. In addition, by exploiting the redundancy inherent to a signal, CS conducts sampling and compression at the same time, which greatly alleviates the need for high transmission bandwidth and large storage space, enabling low-cost on-sensor data compression. CS has been applied in many practical applications, including but not limited to single-pixel imaging \cite{duarte2008single, rousset2017adaptive}, accelerating magnetic resonance imaging (MRI) \cite{lustig2007sparse}, wireless tele-monitoring \cite{zhang2013compressed} and cognitive radio communication \cite{sharma2016application}.

\begin{figure}[t]
\setlength{\abovecaptionskip}{0.cm}
\setlength{\belowcaptionskip}{1cm}
\centering
\includegraphics[width=0.48\textwidth]{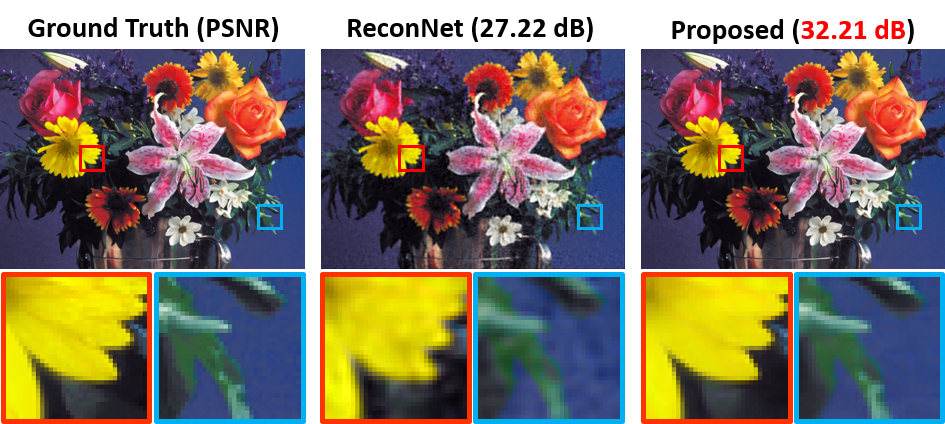}
\caption{CS reconstruction results produced by our proposed ISTA-Net method and a recent network-based CS reconstruction method (ReconNet \cite{kulkarni2016reconnet}), when the CS ratio is 25\%. ISTA-Net clearly produces a higher fidelity reconstruction.}
\vspace{-15pt}
\label{fig: firstfig}
\end{figure}

Mathematically, the purpose of CS reconstruction is to infer the original signal ${\mathbf x}\in \mathbb{R}^N$ from its randomized CS measurements $\mathbf{y = \Phi x} \in \mathbb{R}^M$. Here, $\mathbf{\Phi} \in \mathbb{R}^{M \times N}$ is a linear random projection (matrix). Because $M \ll N$, this inverse problem is typically ill-posed, whereby the CS ratio is defined as $\frac{M}{N}$. In this paper, we mainly focus on CS reconstruction of natural images. However, it is worth noting that our proposed framework can be easily extended to videos and other types of data.

\begin{figure*}[t]
\setlength{\abovecaptionskip}{0.cm}
\setlength{\belowcaptionskip}{-0.1cm}
\centering
\includegraphics[width=0.95\textwidth]{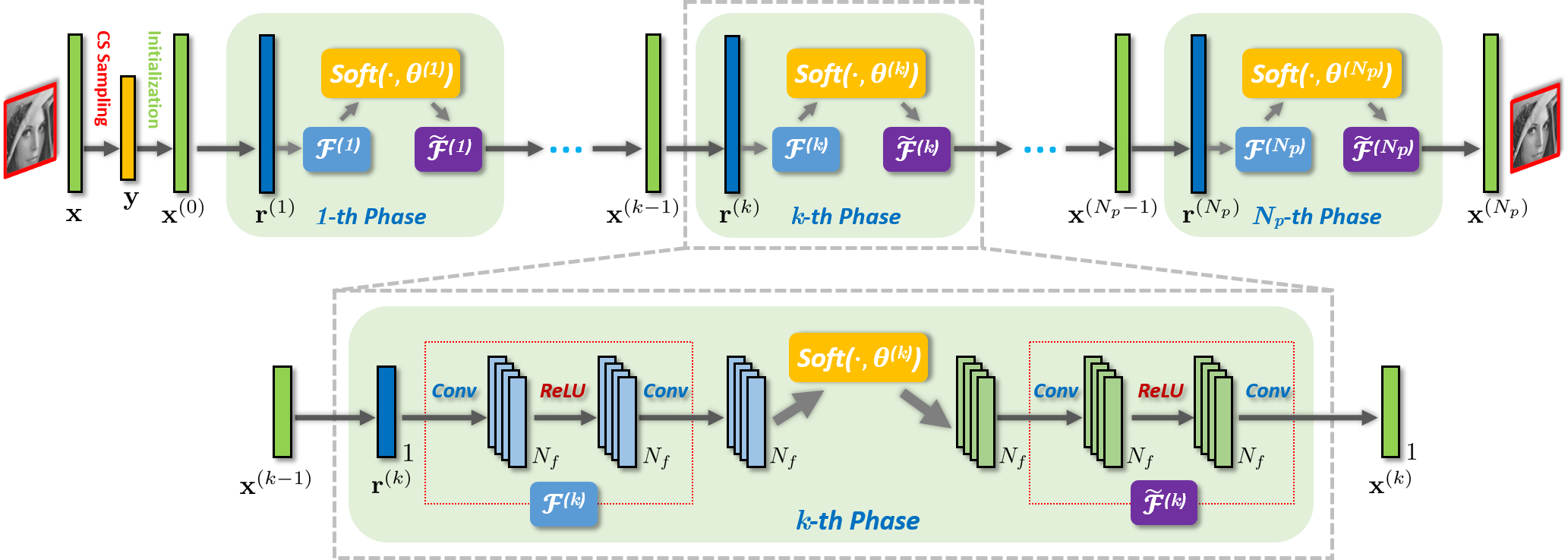}
\caption{\footnotesize Illustration of our proposed ISTA-Net framework. Specifically, ISTA-Net is composed of $N_p$ phases, and each phase strictly corresponds to one iteration in ISTA. The forward transform  $\mathcal{F}^{(k)}$ is designed as a combination of two linear convolutional operators separated by a rectified linear unit (ReLU), and the backward transform $\widetilde{\mathcal{F}}^{(k)}$ is designed to exhibit a structure symmetric to that of $\mathcal{F}^{(k)}$. Note that $\mathcal{F}^{(k)}$ and $\widetilde{\mathcal{F}}^{(k)}$ satisfy the symmetry constraint $\widetilde{\mathcal{F}}^{(k)}\circ \mathcal{F}^{(k)} = \mathcal{I}$, where $ \mathcal{I}$ is the identity operator. $N_f$ denotes the number of feature maps.} \vspace{-14pt}
\label{fig: istanet}
\end{figure*}

In the past decade, a great deal of image CS reconstruction methods have been developed \cite{mun2009block, dong2014compressive, iliadis2018deep, kulkarni2016reconnet}. Most traditional methods exploit some structured sparsity as an  image prior and then solve a sparsity-regularized optimization problem in an iterative fashion \cite{kim2010compressed, li2013efficient, GSR2014, zhang2014image, metzler2016denoising}. Based on some well-studied image formation models and  intrinsic image properties, these methods enjoy the advantages of strong convergence and theoretical analysis in most cases. However, they usually suffer from high computational complexity, and they are also faced with the challenges of choosing optimal transforms and tuning parameters in their solvers. Fueled by the powerful learning ability of deep networks, several deep network-based image CS reconstruction algorithms have been recently proposed to directly learn the inverse mapping from the CS measurement domain to the original signal domain \cite{mousavi2015deep, iliadis2018deep}. Compared to optimization-based algorithms, these non-iterative algorithms dramatically reduce time complexity, while achieving impressive reconstruction performance. However, existing network-based CS algorithms are trained as a \emph{black box}, with limited insights from the CS domain.

In this paper, we design a novel deep architecture, dubbed ISTA-Net, by mapping the traditional ISTA \cite{beck2009fast} for optimizing a general $\ell_1$ norm CS reconstruction model into a deep network. In particular, ISTA-Net is composed of a fixed number of phases, each of which strictly corresponds to an ISTA-like iteration, as illustrated in Figure~\ref{fig: istanet}. Rather than traditional linear transforms, nonlinear learnable and sparsifying transforms are adopted in ISTA-Net, and an efficient and effective strategy to solve the proximal mapping of the nonlinear transform is developed. All the parameters involved in ISTA-Net (\eg nonlinear transforms, shrinkage thresholds, step sizes, etc.) are learned end-to-end using  back-propagation. Moreover, borrowing more insights from the compression realm, an enhanced version, dubbed {ISTA-Net}$^+$, is derived from ISTA-Net by sparsifying natural images explicitly in the residual domain. Interestingly, the skip connections introduced by {ISTA-Net}$^+$ further facilitate the network training. In fact, the proposed ISTA-Nets can be viewed as an attempt to bridge the gap between the two aforementioned categories of CS methods.


\noindent\textbf{Contributions.} The main contributions of this paper are summarized as follows: \textbf{1)} We develop a novel ISTA-Net, which adopts the structure of ISTA update steps to design a learnable deep network manifestation, where all  parameters are discriminately learned instead of being hand-crafted or fixed. By learning sparsifying transforms in the residual domain, an enhanced version {ISTA-Net}$^+$ is derived to further improve CS performance. As such, ISTA-Nets enjoy the advantages of  fast and accurate reconstruction with well-defined interpretability. \textbf{2)} The proximal mapping problem associated to a nonlinear sparsifying transform is solved in a general and efficient way, which actually enables mapping other optimization algorithms into deep network form.  \textbf{3)} Extensive experiments on natural image and MRI CS reconstruction clearly show that ISTA-Net significantly outperforms the state-of-the-art, while maintaining attractive computational complexity.


\section{Related Work}
We generally group  existing CS reconstruction methods of images into two categories: \textit{traditional optimization-based} CS methods and recent \textit{network-based} CS methods. In what follows, we give a brief review of both and focus on the specific methods most relevant to our own.

\vspace{3pt}\noindent{\textbf{Optimization-based CS reconstruction:}} Given the linear measurements $\mathbf{y}$, traditional image CS methods usually reconstruct the original image $\mathbf{x}$ by solving the following (generally convex) optimization problem:
\begin{equation}
\setlength{\abovedisplayskip}{5pt}
\setlength{\belowdisplayskip}{5pt}
\underset{\mathbf x}{\min} ~~\frac{1}{2}\|\mathbf{\Phi x - y}\|^2_2 + \lambda \|\mathbf{\Psi x}\|_1, \label{eq: classical CS}
\end{equation}
\noindent where $\mathbf{\Psi x}$ denotes the transform coefficients of  $\mathbf{x}$ with respect to some transform $\mathbf{\Psi}$ and the sparsity of the vector $\mathbf{\Psi x}$ is encouraged  by the $\ell_1$ norm with 
$\lambda$ being the (generally pre-defined) regularization parameter.

Since natural images are typically non-stationary, the classic fixed domains (\eg DCT, wavelet \cite{mun2009block}, and gradient domain \cite{li2013efficient}) usually result in poor reconstruction performance. Many works incorporate additional prior knowledge about transform coefficients (\eg statistical dependencies \cite{kim2010compressed, zhao2017video}, structure \cite{he2009exploiting}, etc.) into the CS reconstruction framework. Furthermore, some elaborate priors exploiting the non-local self-similarity properties of natural images have been proposed to improve CS reconstruction \cite{RCoS2012, GSR2014, dong2014compressive, zhao2016nonconvex}.  Metzler \etal propose to integrate the well-defined BM3D denoiser into the approximate message passing (AMP) framework for CS reconstruction \cite{metzler2016denoising}. Quite recently, some fast and effective convolutional neural network (CNN) denoisers are trained and integrated into the half quadratic splitting (HQS) \cite{zhang2017learning} and the alternating direction method of multipliers (ADMM) \cite{chang2017projector} to solve image inverse problems. However, all these traditional image CS methods require hundreds of iterations to solve Eq.~(\ref{eq: classical CS}) by means of various iterative solvers, which inevitably gives rise to high computational cost thus restricting the application of CS. In addition, the selected image prior (\eg optimal transform) or the optimization parameters (\eg step size and regularization parameter) are usually hand-crafted and quite challenging to pre-define.

\vspace{3pt}\noindent {\textbf{Network-based CS reconstruction:}} Inspired by the powerful learning capability of deep networks \cite{xie2012image} and its success in computer vision tasks  \cite{krizhevsky2012imagenet,long2015fully}, several deep network-based image CS reconstruction algorithms have recently been  proposed \cite{mousavi2015deep,iliadis2018deep,kulkarni2016reconnet,mousavi2017learning}. Mousavi \etal first propose to apply a stacked denoising auto-encoder (SDA) to learn the representation from training data and to reconstruct test data from their CS measurements \cite{mousavi2015deep}. Adler \etal and Iliadis \etal separately propose to utilize fully-connected neural networks for image and video CS reconstruction \cite{adler2016deep, iliadis2018deep}. Kulkarni \etal further develop a CNN-based CS algorithm, dubbed ReconNet, which learns to regress an image block (output) from its  CS measurement (input) \cite{kulkarni2016reconnet}. Mousavi and Baraniuk recently propose an all-convolutional network for image CS reconstruction \cite{mousavi2017learning}. A main feature of network-based image CS methods is that they are non-iterative, which dramatically reduces time complexity as compared with their optimization-based counterparts. However, this is done with either  fully-connected or repetitive convolutional layers. We believe that their lack of structural diversity, which originates from the absence of CS domain specific insights inherent to optimization-based methods, is the bottleneck for further performance improvement.

The tremendous success of deep learning for many image processing applications has also led researchers to consider relating iterative optimization methods to neural networks \cite{jin2017deep, wang2016proximal, xin2016maximal, riegler2016atgv}. For instance, in the context of sparse coding, Grefor and LeCun propose a fast algorithm to calculate good approximations of optimal sparse codes by introducing the Learned ISTA (LISTA), in which two matrices in classical ISTA are learned instead of using pre-computed ones \cite{gregor2010learning}. Mark \etal extend  approximate message passing (AMP) algorithms to so-called Learned AMP networks for solving sparse linear inverse problems \cite{borgerding2017amp}. Relying on LISTA, some sparse-coding based networks for image super-resolution and deblocking are proposed \cite{wang2015deep, wang2016d3, liu2016robust}.
For image denoising and deconvolution, Schmidt and Roth propose to learn the linear filters and shrinkage functions under the framework of half-quadratic optimization \cite{schmidt2014shrinkage}. Chen \etal propose a trainable reaction diffusion model by learning several parameterized linear filters and influence functions for image denoising and deblocking \cite{chen2015learning}. In the context of CS for sparse signals, Kamilov and Mansour propose to learn the optimal thresholding functions for ISTA based on a B-spline decomposition \cite{kamilov2016learning}.

Recently, Yang \etal propose a so-called ADMM-Net architecture by reformulating ADMM for CS magnetic resonance imaging (CS-MRI) using deep networks \cite{yang2016deep}. Although both ADMM-Net and our proposed ISTA-Net have similar inspirations, they are quite different. In fact, there are two main differences between both methods. First, ADMM-Net is specifically designed and developed for CS-MRI based on ADMM, while our ISTA-Net is much more general, since it works well for both general CS and CS-MRI based on ISTA. Second, ADMM-Net only utilizes several linear filters, while ISTA-Net goes beyond that to adopt nonlinear transforms to more effectively sparsify natural images and develops an efficient strategy for solving their proximal mapping problems. The detailed comparison with ADMM-Net for CS-MRI can be found in Section~ \ref{subsection:ADMM}.

In a nutshell, the proposed ISTA-Net can be  essentially viewed as a significant extension of  LISTA \cite{gregor2010learning}, from the sparse coding problem to general CS reconstruction. Compared with traditional optimization-based CS methods, ISTA-Net is able to learn its optimal parameters, \ie thresholds, step sizes as well as nonlinear transforms, without hand-crafted settings. In addition, ISTA-Net  has the same computational complexity as several iterations of traditional ISTA, which is more than 100 times faster than existing methods of this category. Compared with network-based CS methods, ISTA-Net borrows insights from traditional optimization methods to allow for interpretability in its network design and it utilizes the  structural diversity originating from the CS domain. Extensive experiments demonstrate that ISTA-Net significantly outperforms the existing optimization-based and network-based CS methods, even when compared against methods that are designed for a specific domain (\eg CS-MRI).

\section{Proposed ISTA-Net for Compressive Sensing}
In this section, we first briefly review traditional ISTA optimization for image CS reconstruction, and then elaborate on the design of our proposed ISTA-Net.

\subsection{Traditional ISTA for CS}
The iterative shrinkage-thresholding algorithm (ISTA) is a popular first order proximal method, which is well suited for solving many large-scale linear inverse problems. Specifically, ISTA solves the CS reconstruction problem in Eq.~(\ref{eq: classical CS}) by iterating between the following update steps:
\begin{equation}
\setlength{\abovedisplayskip}{7pt}
\setlength{\belowdisplayskip}{5pt}
\mathbf{r}^{(k)}= \mathbf{x}^{(k-1)} - \rho \mathbf{\Phi}^{\top} (\mathbf{\Phi} \mathbf{x}^{(k-1)} - \mathbf{y}), \label{eq: update r ISTA}
\end{equation}
\begin{equation}
\setlength{\abovedisplayskip}{5pt}
\setlength{\belowdisplayskip}{5pt}
\mathbf{x}^{(k)} = \underset{\mathbf x}{\arg\min} ~~\frac{1}{2}\|\mathbf{x} - \mathbf{r}^{(k)}\|^2_2  + \lambda \|\mathbf{\Psi x}\|_1. \label{eq: update x ISTA}
\end{equation}
Here, $k$ denotes the ISTA iteration index, and $\rho$ is the step size. Eq.~(\ref{eq: update r ISTA}) is trivial, while Eq.~(\ref{eq: update x ISTA}) is actually a special case of the so-called proximal mapping, \ie $\mathtt{prox}_{\lambda\phi}(\mathbf{r}^{(k)})$, when $\phi(\mathbf{x})= \|\mathbf{\Psi x}\|_1$. Formally, the proximal mapping of regularizer $\phi$ denoted by $\mathtt{prox}_{\lambda\phi}(\mathbf{r})$ is defined as
\begin{equation}
\setlength{\abovedisplayskip}{5pt}
\setlength{\belowdisplayskip}{5pt}
\mathtt{prox}_{\lambda\phi}(\mathbf{r})=\arg\min_\mathbf{x}~~\frac{1}{2}||\mathbf{x}-\mathbf{r}||^2_2+\lambda\phi(\mathbf{x}).
\end{equation}

Solving $\mathtt{prox}_{\lambda\phi}(\mathbf{r})$ in an efficient and effective way is critical for ISTA \cite{zhang2013structural}, as well as for other optimization methods, such as ADMM \cite{afonso2011augmented} and AMP \cite{metzler2016denoising}. For example, when  $\mathbf{\phi(x)}=\|\mathbf{W x}\|_1$ ($\mathbf{W}$ is wavelet transform matrix), we have $\mathtt{prox}_{\lambda\phi}(\mathbf{r}) = \mathbf{W}^{\top} soft(\mathbf{W}\mathbf{r}, \lambda)$ due to the orthogonality of $\mathbf{W}$. However, it remains non-trivial to obtain $\mathbf{x}^{(k)}$ in Eq.~(\ref{eq: update x ISTA}) for a more complex non-orthogonal (or even nonlinear) transform $\mathbf{\Psi}$. In addition, ISTA usually requires many iterations to obtain a satisfactory result, suffering from extensive computation. The optimal transform $\mathbf{\Psi}$ and all the parameters such as $\rho$ and $\lambda$ are pre-defined (do not change with $k$), and very challenging to tune apriori. 

\subsection{ISTA-Net Framework}
\label{subsec: ISTA-Net}
By taking full advantage of the merits of ISTA-based and network-based CS methods, the basic idea of ISTA-Net is to map the previous ISTA update steps to a deep network architecture that consists of  a fixed number of phases, each of which corresponds to one iteration in traditional ISTA.

In order to improve  reconstruction performance and increase network capacity and instead of the hand-crafted transform $\mathbf{\Psi}$ in Eq.~(\ref{eq: classical CS}), ISTA-Net adopts a general nonlinear transform function to sparsify natural images, denoted by $\mathcal{F}(\cdot)$, whose parameters are learnable. In particular and inspired by the powerful representation power of CNN  \cite{dong2014learning} and its universal approximation property \cite{hornik1989multilayer}, we propose to design $\mathcal{F}(\cdot)$ as a combination of two linear convolutional operators (without bias terms) separated by a rectified linear unit (ReLU). As illustrated in Figure~\ref{fig: istanet}, the first convolutional operator in $\mathcal{F}(\cdot)$ corresponds to $N_f$ filters (each of size  $3\times 3$ in our experiments) and the second convolutional operator corresponds to another set of $N_f$ filters (each of size  $3\times 3 \times N_f$ in our experiments). In our implementation, we set $N_f=32$ by default. Obviously, $\mathcal{F}(\cdot)$ can also be equivalently formulated in matrix form as  $\mathcal{F}(\mathbf{x}) = \mathbf{B}{ReLU}(\mathbf{A}\mathbf{x})$, where $\mathbf{A}$ and $\mathbf{B}$ correspond to the above two convolutional operators, respectively. With its learnable and nonlinear characteristics, $\mathcal{F}(\cdot)$ is expected to be able to achieve a richer representation for natural images.

Replacing $\mathbf{\Psi}$ in Eq.~(\ref{eq: classical CS}) with $\mathcal{F}(\cdot)$, we obtain the following sparsity-inducing regularization problem with a nonlinear transform:
\begin{equation}
\setlength{\abovedisplayskip}{5pt}
\setlength{\belowdisplayskip}{5pt}
\underset{\mathbf x}{\min} ~~\frac{1}{2}\|\mathbf{\Phi x - y}\|^2_2 + \lambda \|\mathcal{F}(\mathbf{x})\|_1. \label{eq: ISTA-Net CS}
\end{equation}

By solving Eq.~(\ref{eq: ISTA-Net CS}) using ISTA, Eq.~(\ref{eq: update r ISTA}) is unchanged while Eq.~(\ref{eq: update x ISTA}) becomes
\begin{equation} \label{eq: ISTAnet x update}
\setlength{\abovedisplayskip}{5pt}
\setlength{\belowdisplayskip}{5pt}
\mathbf{x}^{(k)} =  \underset{\mathbf x}{\arg\min} ~~\frac{1}{2}\|\mathbf{x} - \mathbf{r}^{(k)}\|^2_2  + \lambda \|\mathcal{F}(\mathbf{x})\|_1.
\end{equation}

In the following, we argue that the above two steps Eq.~(\ref{eq: update r ISTA}) and Eq.~(\ref{eq: ISTAnet x update}) in  the $k$-\textit{th}  ISTA iteration  both admit efficient solutions, and we cast them into two separate modules in the $k$-\textit{th} phase of ISTA-Net, namely the $\mathbf{r}^{{(k)}}$ {\textbf{module}} and the $\mathbf{x}^{{(k)}}$ {\textbf{module}}, as illustrated in Figure~\ref{fig: istanet}. 

\vspace{3pt}\noindent  $\bullet~~\mathbf{r}^{{(k)}}$ \textbf{Module:} It corresponds to Eq.~(\ref{eq: update r ISTA}) and is used to generate the immediate reconstruction result $\mathbf{r}^{(k)}$. Note that $\mathbf{\Phi}^{\top} (\mathbf{\Phi} \mathbf{x}^{(k-1)} - \mathbf{y})$ is essentially the gradient of the data-fidelity term $\frac{1}{2}\|\mathbf{\Phi} \mathbf{x} - \mathbf{y}\|^2_2$, computed at $\mathbf{x}^{(k-1)}$. To preserve the ISTA structure while increasing network flexibility, we allow the step size $\rho$ to vary across  iterations (while it is fixed in traditional ISTA), so the output of this module with input $\mathbf{x}^{(k-1)}$ is finally defined as:
\begin{equation} \label{eq: update r module}
\setlength{\abovedisplayskip}{5pt}
\setlength{\belowdisplayskip}{5pt}
\mathbf{r}^{(k)} = \mathbf{x}^{(k-1)} - \rho^{(k)} \mathbf{\Phi}^{\top} (\mathbf{\Phi} \mathbf{x}^{(k-1)} - \mathbf{y}).
\end{equation}

\vspace{3pt}\noindent $\bullet~~\mathbf{x}^{{(k)}}$ \textbf{Module:} It aims to compute $\mathbf{x}^{(k)}$ according to Eq.~(\ref{eq: ISTAnet x update}) with  input $\mathbf{r}^{(k)}$. Note that Eq.~(\ref{eq: ISTAnet x update}) is actually the proximal mapping $\mathtt{prox}_{\lambda \mathcal{F}}(\mathbf{r}^{(k)})$ associated with the nonlinear transform $\mathcal{F}(\cdot)$. In this paper, we propose to solve $\mathtt{prox}_{\lambda \mathcal{F}}(\mathbf{r}^{(k)})$ efficiently in two steps, which is also one of our main contributions.

First, note that $\mathbf{r}^{(k)}$ is the immediate reconstruction result of $\mathbf{x}^{(k)}$ at the $k$-\textit{th} iteration. In the context of image inverse problems, one general and reasonable assumption is that each element of $(\mathbf{x}^{(k)} - \mathbf{r}^{(k)})$ follows an independent normal distribution with common zero mean and variance $\sigma^2$ \cite{GSR2014}. Here, we also make this assumption, and then we further prove the following theorem:

\begin{theorem}\label{theorem: F module}
Let $X_1, ...,X_n$ be independent normal random variables with common zero mean and variance $\sigma^2$. If $\vec{X}=[X_1, ..., X_n]^{\top}$ and given any matrices $\mathbf{A} \in \mathbb{R}^{m \times n}$ and $\mathbf{B} \in \mathbb{R}^{s \times m}$, define a new random variable $\vec{Y} = \mathbf{B}{ReLU}(\mathbf{A} \vec{X}) = \mathbf{B} \max(\mathbf{0}, \mathbf{A} \vec{X})$. Then, $ \mathbb{E}[\|\vec{Y} - \mathbb{E}[\vec{Y}]\|^2_2] $ and $\mathbb{E}[\|\vec{X} - \mathbb{E}[\vec{X}]\|^2_2] $ are linearly related, \ie $ \mathbb{E}[\|\vec{Y} - \mathbb{E}[\vec{Y}]\|^2_2] = \alpha \mathbb{E}[\|\vec{X} - \mathbb{E}[\vec{X}]\|^2_2] $, where $\alpha$ is only a function of  $\mathbf{A}$ and $\mathbf{B}$. (Please refer to the \textbf{supplementary material} for the proof and more details.)
\end{theorem}

\begin{figure*}[t]
\setlength{\abovecaptionskip}{0.cm}
\setlength{\belowcaptionskip}{-0.3cm}
\centering
\includegraphics[width=0.85\textwidth]{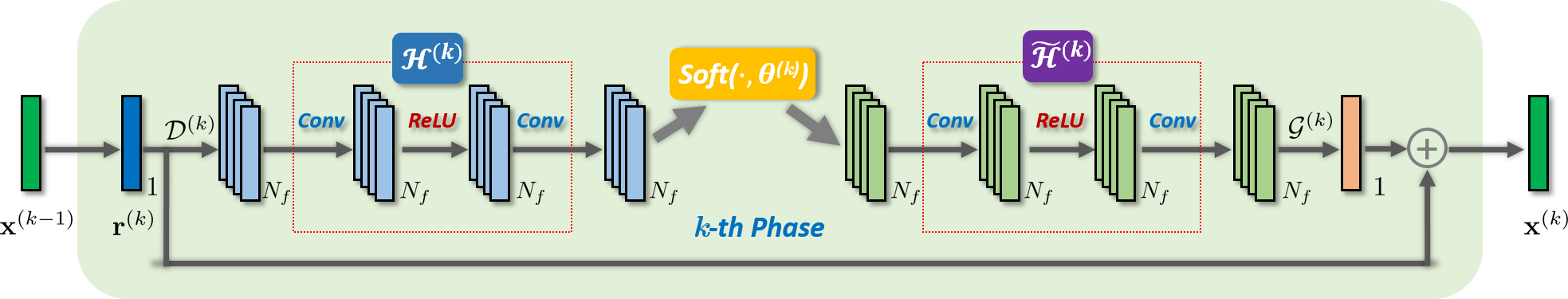}
\caption{Illustration of the $k$-\textit{th} phase of the proposed ISTA-Net$^+$. $\mathcal{D}^{(k)}, \mathcal{G}^{(k)}, \mathcal{H}^{(k)}, \widetilde{\mathcal{H}}^{(k)}$ are learnable linear convolutional operators.} \label{fig: ista-net-plus_phase}
\vspace{-12pt}
\end{figure*}

\textbf{Theorem} \ref{theorem: F module} can be easily extended to a normal distribution. Suppose that $\mathbf{r}^{(k)}$ and $\mathcal{F}(\mathbf{r}^{(k)})$ are the mean values of $\mathbf{x}$ and $\mathcal{F}(\mathbf{x})$ respectively, then we can make the following approximation based on \textbf{Theorem} \ref{theorem: F module}:
\begin{equation}
\setlength{\abovedisplayskip}{5pt}
\setlength{\belowdisplayskip}{5pt}
\|\mathcal{F}(\mathbf{x})-\mathcal{F}(\mathbf{r}^{(k)})\|^2_2 \approx \alpha \|\mathbf{x}-\mathbf{r}^{(k)}\|^2_2,
\end{equation}
where $\alpha$ is a scalar that is only related to the parameters of $\mathcal{F}(\cdot)$. By incorporating this linear relationship into Eq. (\ref{eq: ISTAnet x update}), we obtain the following optimization:
\begin{equation}
\setlength{\abovedisplayskip}{5pt}
\setlength{\belowdisplayskip}{5pt}
\text{\small $\mathbf{x}^{(k)} =  \underset{\mathbf x}{\arg\min} ~~\frac{1}{2}\|\mathcal{F}(\mathbf{x}) - \mathcal{F}(\mathbf{r}^{(k)})\|^2_2  +  \theta \|\mathcal{F}(\mathbf{x})\|_1,\label{eq: F(x)}$}
\end{equation}
where $\lambda$ and $\alpha$ are merged into one parameter $\theta$, \ie $\theta = \lambda \alpha$. Therefore, we get a closed-form version of $\mathcal{F}(\mathbf{x}^{(k)})$:
\begin{equation}
\setlength{\abovedisplayskip}{5pt}
\setlength{\belowdisplayskip}{5pt}
\mathcal{F}(\mathbf{x}^{(k)}) = soft(\mathcal{F}(\mathbf{r}^{(k)}), \theta ).
\end{equation}

Second, motivated by the invertible characteristics of the wavelet transform that leads to a closed-form solution for Eq.~(\ref{eq: update x ISTA}), we  introduce the left inverse of ${\mathcal{F}}(\cdot)$, denoted by  $\widetilde{\mathcal{F}}(\cdot)$ such that $\widetilde{\mathcal{F}} \circ \mathcal{F} = \mathcal{I}$, where $\mathcal{I}$ is the identity operator. Specifically, $\widetilde{\mathcal{F}}(\cdot)$ is designed to exhibit a  structure symmetric to that of $\mathcal{F}(\cdot)$, so it is also modeled as two linear convolutional operators separated by a ReLU operator, as shown in Figure~\ref{fig: istanet}. 
Because $\mathcal{F}(\cdot)$ and $\widetilde{\mathcal{F}}(\cdot)$ are both learnable, we will enforce the \textbf{\textit{{symmetry constraint}}} $\widetilde{\mathcal{F}} \circ \mathcal{F} = \mathcal{I}$  by incorporating it into the loss function during network training. 
Therefore, 
$\mathbf{x}^{(k)}$ can be efficiently computed in closed-form as:
\begin{equation}
\setlength{\abovedisplayskip}{5pt}
\setlength{\belowdisplayskip}{5pt}
\mathbf{x}^{(k)} = \widetilde{\mathcal{F}}(soft(\mathcal{F}(\mathbf{r}^{(k)}), \theta )).
\label{eq: update x solution}
\end{equation}

It is worth emphasizing that $\theta$, as a shrinkage threshold, is a learnable parameter in this module. Similarly, to increase network capacity, we do not constrain that $\mathcal{F}(\cdot)$, $\widetilde{\mathcal{F}}(\cdot)$, and $\theta$ be the same at each phase. That is, each phase of ISTA-Net has its own $\{\mathcal{F}^{(k)}(\cdot),\widetilde{\mathcal{F}}^{(k)}(\cdot),\theta^{(k)}\}$, as illustrated in Figure~\ref{fig: istanet}. Therefore, with all the learnable parameters, the output $\mathbf{x}^{(k)}$ in this module should be updated as:
\begin{equation} \label{eq: update x module}
\setlength{\abovedisplayskip}{5pt}
\setlength{\belowdisplayskip}{5pt}
\mathbf{x}^{(k)} = \widetilde{\mathcal{F}}^{(k)}(soft(\mathcal{F}^{(k)}(\mathbf{r}^{(k)}), \theta^{(k)} )).
\end{equation}

Figure~\ref{fig: istanet} clearly illustrates how Eq.~(\ref{eq: ISTAnet x update}) with the closed-form solution in Eq.~(\ref{eq: update x module}) is mapped into a deep network  in the $k$-\textit{th} phase of ISTA-Net.

\vspace{3pt}\noindent \textbf{Parameters in ISTA-Net:}
Each module in each phase of ISTA-Net strictly corresponds to the updates steps in an ISTA iteration. The learnable parameter set in ISTA-Net, denoted by $\mathbf{\Theta}$, includes the step size $\rho^{(k)}$ in the $\mathbf{r}^{(k)}$ module, the parameters of the forward and backward transforms $\mathcal{F}^{(k)}(\cdot)$ and $\widetilde{\mathcal{F}}^{(k)}(\cdot)$, and the shrinkage threshold $\theta^{(k)}$ in the $\mathbf{x}^{(k)}$ module. As such,  $\mathbf{\Theta} = \{\rho^{(k)}, \theta^{(k)}, \mathcal{F}^{(k)}, \widetilde{\mathcal{F}}^{(k)}\}_{k=1}^{N_p}$, where $N_p$ is the total number of ISTA-Net phases. All these parameters will be learned as neural network parameters.

\vspace{3pt}\noindent \textbf{Initialization:}
Like traditional ISTA, ISTA-Net also requires an initialization denoted by $\mathbf{x}^{(0)}$ in Figure~\ref{fig: istanet}. Instead of random values, we propose to directly use a linear mapping to compute the initialization. Specifically, given the training data pairs that include the image blocks and their corresponding CS measurements, \ie $\left \{(\mathbf{y}_i, \mathbf{x}_i) \right \}_{i=1}^{N_b}$ with $\mathbf{x}_i\in\mathbb{R}^N, \mathbf{y}_i\in\mathbb{R}^M$, the linear mapping matrix, denoted by $\mathbf{Q}_{init}$, can be computed by solving a least squares problem: $\mathbf{Q}_{init} = \arg\min_{\mathbf{Q}}  \|\mathbf{Q}\mathbf{Y} - \mathbf{X}\|^2_F=\mathbf{X}\mathbf{Y}^{\top}(\mathbf{Y}\mathbf{Y}^{\top})^{-1}$. Here, $\mathbf{X} = [\mathbf{x}_1, ..., \mathbf{x}_{N_b}]$, and $\mathbf{Y} = [\mathbf{y}_1, ..., \mathbf{y}_{N_b}]$. 
Hence, given any input CS measurement $\mathbf{y}$, its corresponding ISTA-Net initialization $\mathbf{x}^{(0)}$ is computed as: $\mathbf{x}^{(0)} = \mathbf{Q}_{init}\mathbf{y}.$

\subsection{Loss Function Design}
\label{LossFunctionDesign}
Given the training data pairs $\left \{ (\mathbf{y}_i, \mathbf{x}_i) \right \}_{i=1}^{N_b}$, ISTA-Net first takes the CS measurement $\mathbf{y}_i$ as input and generates the reconstruction result, denoted by $\mathbf{x}_i^{(N_p)}$, as output. We seek to reduce the discrepancy between $\mathbf{x}_i$ and $\mathbf{x}^{(N_p)}_i$ while satisfying the symmetry constraint $\widetilde{\mathcal{F}}^{(k)}\circ \mathcal{F}^{(k)}=\mathcal{I}~~ \forall k=1, \ldots, N_p$. Therefore, we design the end-to-end loss function for ISTA-Net as follows:
\begin{align}
&{\mathcal{L}}_{total}(\mathbf{\Theta}) =  {\mathcal{L}}_{discrepancy}+\gamma{\mathcal{L}}_{constraint}, \label{eq: LossFunction}\\
&\text{with:}
\begin{cases}
\text{\small ${\mathcal{L}}_{discrepancy}=\frac{1}{N_bN}{\sum^{N_b}_{i=1}\|\mathbf{x}^{(N_p)}_i - \mathbf{x}_i\|^2_2}$}\\
\text{\small ${\mathcal{L}}_{constraint}=\frac{1}{N_bN}{\sum^{N_b}_{i=1}\sum^{N_p}_{k=1}\|\widetilde{\mathcal{F}}^{(k)}(\mathcal{F}^{(k)}(\mathbf{x}_i)) - \mathbf{x}_i\|^2_2}$},
\end{cases}\notag
\end{align}
where $N_p$, $N_b$, $N$, and $\gamma$ are the total number of ISTA-Net phases, the total number of training blocks, the size of each block $\mathbf{x}_i$, and the regularization parameter, respectively. In our experiments, $\gamma$ is set to 0.01.

\section{Enhanced Version: ISTA-Net$^+$}

Motivated by the fact that the residuals of natural images and videos are more compressible \cite{wallace1992jpeg, sullivan2012overview}, an enhanced version, dubbed ISTA-Net$^+$, is derived from ISTA-Net to further improve CS performance. Starting from Eq.~(\ref{eq: ISTAnet x update}), we assume that $\mathbf{x}^{(k)}=\mathbf{r}^{(k)}+ \mathbf{w}^{(k)}+\mathbf{e}^{(k)}$, where $\mathbf{e}^{(k)}$ stands for some noise and $\mathbf{w}^{(k)}$ represents some missing high-frequency component in $\mathbf{r}^{(k)}$, which can be extracted by a linear operator $\mathcal{R}(\cdot)$ from $\mathbf{x}^{(k)}$, \ie $\mathbf{w}^{(k)}= \mathcal{R}(\mathbf{x}^{(k)})$. Furthermore, $\mathcal{R}(\cdot)$ is defined as  $\mathcal{R}=\mathcal{G} \circ \mathcal{D}$, where $\mathcal{D}$ corresponds to $N_f$ filters (each of size $3\times 3$ in our experiments) and $\mathcal{G}$ corresponds to $1$ filter (with size  $3\times 3 \times N_f$). By modeling $\mathcal{F}=\mathcal{H} \circ \mathcal{D}$, where $\mathcal{H}$ consists of two linear convolutional operators and one ReLU, as illustrated in Figure~\ref{fig: ista-net-plus_phase}, we can replace  $\mathcal{F}$ in Eq.~(\ref{eq: F(x)}) with $\mathcal{H} \circ \mathcal{D}$ to obtain:
\begin{equation}\label{eq: ista_net_plus_F(x)}
\setlength{\abovedisplayskip}{5pt}
\setlength{\belowdisplayskip}{5pt}
\text{\small $\min_\mathbf{x}~~\frac{1}{2}|| {\mathcal{H}(\mathcal{D}(\mathbf x))-\mathcal{H}(\mathcal{D}(\mathbf{r}^{(k)}))} ||_2^2+\theta ||\mathcal{H}(\mathcal{D}(\mathbf{x}))||_1$}.
\end{equation}

\vspace{3pt} By exploiting the approximation used in Eq.~(\ref{eq: F(x)}) and following the same strategy as in ISTA-Net, we define the left inverse of ${\mathcal{H}}$ as $\widetilde{\mathcal{H}}$, which has a structure symmetric to that of ${\mathcal{H}}$ and satisfies the symmetry constraint $\widetilde{\mathcal{H}} \circ {\mathcal{H}} = {\mathcal{I}}$. Thus, the closed form of the ISTA-Net$^+$ update for $\mathbf{x}^{(k)}$ is:
\begin{equation}\label{eq: ista_net_plus_x}
\setlength{\abovedisplayskip}{5pt}
\setlength{\belowdisplayskip}{5pt}
\mathbf{x}^{(k)}=\mathbf{r}^{(k)}+ \mathcal{G}(\widetilde{\mathcal{H}}(soft(\mathcal{H}(\mathcal{D}(\mathbf{r}^{(k)})), \theta))).
\end{equation}

\vspace{3pt} Similar to ISTA-Net, each phase of ISTA-Net$^+$ also has its own learnable parameters, and the $k$-\textit{th} phase of ISTA-Net$^+$ is illustrated in Figure~\ref{fig: ista-net-plus_phase}. Hence, the learnable parameter set $\mathbf{\Theta}^+$ of ISTA-Net$^+$ is $\mathbf{\Theta}^+=\{\rho^{(k)}, \theta^{(k)}, \mathcal{D}^{(k)}, \mathcal{G}^{(k)}, \mathcal{H}^{(k)}, \widetilde{\mathcal{H}}^{(k)} \}_{k=1}^{N_p}$. The loss function of ISTA-Net$^+$ is analogously designed by incorporating the constraints $\widetilde{\mathcal{H}}^{(k)}\circ \mathcal{H}^{(k)}=\mathcal{I}$ into Eq.~(\ref{eq: LossFunction}).

\begin{figure}[t]
\setlength{\abovecaptionskip}{0.cm}
\setlength{\belowcaptionskip}{-0.3cm}
\centering
\includegraphics[width=0.4\textwidth]{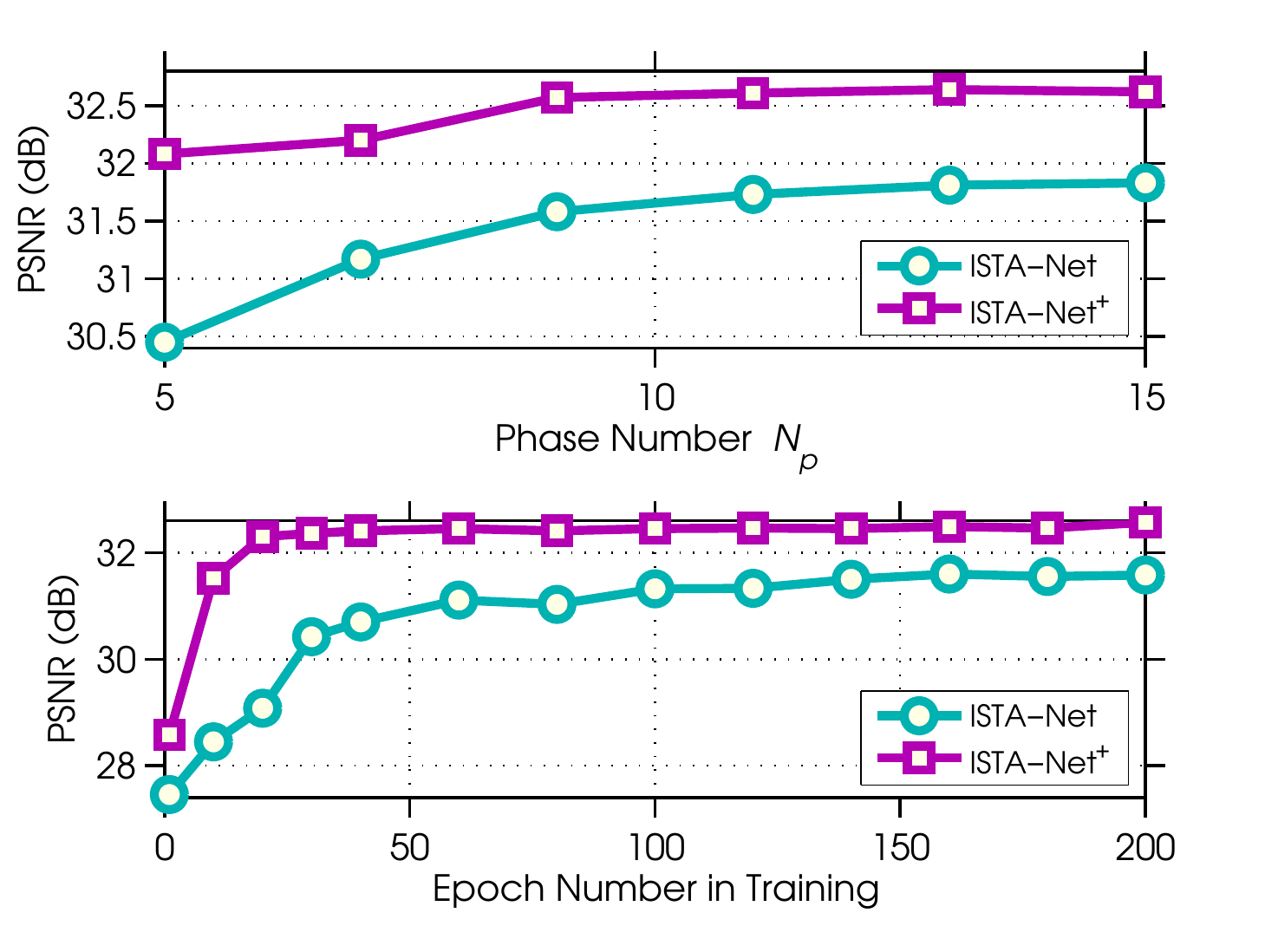}
\caption{PSNR comparison between ISTA-Net and ISTA-Net$^+$ with various numbers of phases and epochs during training.}
\label{fig: ISTA-Net vs ISTA-Net+}
\vspace{-16pt}
\end{figure}

\begin{figure*}[t]
\setlength{\abovecaptionskip}{0cm}
\setlength{\belowcaptionskip}{-0.5cm}
\centering
\includegraphics[width=0.92\textwidth]{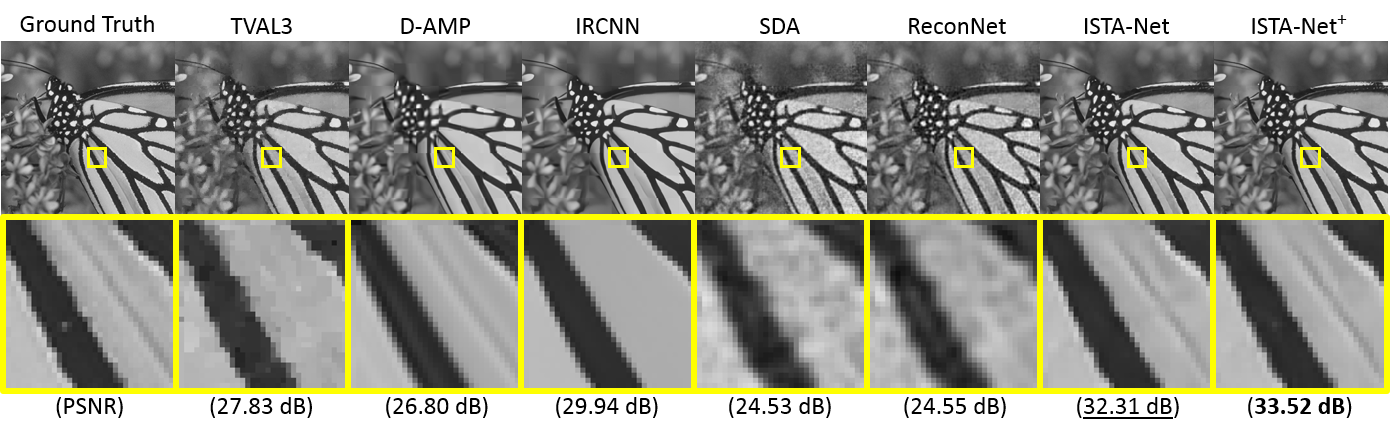}
\caption{Comparison of seven CS reconstruction methods (including our ISTA-Net and ISTA-Net$^+$), when applied to the \textit{Butterfly} image in Set11 (CS ratio is 25\%).}
\label{fig: visual}
\end{figure*}

\begin{table*}[t]
\centering
\caption{Average PSNR (dB) performance comparisons on Set11 with different CS ratios. The best performance is labeled in bold and the second best is underlined. Note that the last two columns is a run-time analysis of all the competing methods, showing the average time to reconstruct a $256\times256$ image and the corresponding frames-per-second (FPS).}
\small
\begin{tabular}{|l|c|c|c|c|c|c|c||c|c|}
\hline
\multirow{2}{*}{\textbf{Algorithm}} & \multicolumn{7}{c||}{\textbf{CS Ratio}}                                 & \multirow{2}{*}{\begin{tabular}[c]{@{}c@{}}\textbf{Time} \\ \small{CPU/GPU} \end{tabular}} & \multicolumn{1}{c|}{\multirow{2}{*}{\begin{tabular}[c]{@{}c@{}}\textbf{FPS} \\ \small{CPU/GPU} \end{tabular}}} \\ \cline{2-8}
                           & 50\%   & 40\%   & 30\%   & 25\%    & 10\%   & 4\%  & 1\%  &                                                                            & \multicolumn{1}{c|}{}                     \\ \hline \hline
TVAL3 \cite{li2013efficient}                     & 33.55 & 31.46 & 29.23 & 27.92  & 22.99 & 18.75 & 16.43 &  3.135s/--------                                                           &  0.32/-----                                     \\
D-AMP \cite{metzler2016denoising}               & 35.92 & 33.56 & 30.39 & 28.46  & 22.64 & 18.40 & 5.21  &   51.21s/--------                                                           & 0.02/-----                                     \\
IRCNN \cite{zhang2017learning}               & 36.23 & 34.06 & 31.18 & 30.07  & 24.02 & 17.56 & 7.70  &     --------/68.42s                                                           & ----/0.015
\\
SDA \cite{mousavi2015deep}                & 28.95  & 27.79  & 26.63  & 25.34   & 22.65 & 20.12 & 17.29 &     ------/{0.0032}s                                                            & ----/{312.5}                                     \\
ReconNet \cite{kulkarni2016reconnet}          & 31.50  & 30.58  & 28.74  & 25.60   & 24.28 & 20.63 & 17.27 &  --------/{0.016}s                                                          & ------/{62.5}                                     \\
                                      \hline \hline
ISTA-Net                                      & \underline{37.43} & \underline{35.36} & \underline{32.91} & \underline{31.53}  & \underline{25.80} & \underline{21.23} & \underline{17.30} & 0.923s/0.039s                                                         & 1.08/25.6                                   \\
ISTA-Net$^+$                      & \textbf{38.07} & \textbf{36.06} & \textbf{33.82} & \textbf{32.57}  & \textbf{26.64} & \textbf{21.31} & \textbf{17.34} & 1.375s/0.047s                                                                     & 0.73/21.3                                    \\ \hline
\end{tabular}
\label{table: all PSNR Set11}
\vspace{-10pt}
\end{table*}

\section{Experimental Results}
For fair comparison, we use the same set of 91 images as in \cite{kulkarni2016reconnet} for training. Following common practices in previous CS work, we  generate the training data pairs $\left \{ (\mathbf{y}_i, \mathbf{x}_i) \right \}_{i=1}^{N_b}$ by first extracting the luminance component of  88,912 randomly cropped image blocks (each of size 33$\times$33), \ie $N_b$=88,912 and $N$=1,089. 
Then, for a given CS ratio, the corresponding measurement matrix $\mathbf{\Phi}\in \mathbb{R}^{M\times N}$ is constructed by generating a random Gaussian matrix and then orthogonalizing its rows, \ie $\mathbf{\Phi}\mathbf{\Phi}^{\top}=\mathbf{I}$, where $\mathbf{I}$ is the identity matrix. Applying $\mathbf{y}_i=\mathbf{\Phi} \mathbf{x}_i$ yields the set of CS measurements, where $\mathbf{x}_i$ is the vectorized version of an image block. We use TensorFlow \cite{abadi2016tensorflow} to implement and train the ISTA-Nets separately for a range of CS ratios $\{1\%, 4\%, 10\%, 25\%, 30\%, 40\%, 50\%\}$. To train the networks, we use Adam optimization \cite{kingma2015adam} with a learning rate of 0.0001 (200 epochs), and a batch size of 64. All the experiments are performed on a workstation with Intel Core i7-6820 CPU and GTX1060 GPU. Training ISTA-Nets with phase number $N_p$=$9$ roughly takes 10 hours. For testing, we utilize two widely used benchmark datasets: Set11 \cite{kulkarni2016reconnet} and BSD68 \cite{martin2001database}, which have 11 and 68 gray images, respectively. The reconstruction results are reported as the average Peak Signal-to-Noise Ratio (PSNR) over the test images.

\subsection{ISTA-Net vs. ISTA-Net$^+$}
To demonstrate the superiority of ISTA-Net$^+$ over ISTA-Net, we compare them in two aspects: performance and convergence. Figure~\ref{fig: ISTA-Net vs ISTA-Net+} (top) shows the average PSNR curves for the  testing set (Set11) with respect to different phase numbers, when the CS ratio is 25\%. We observe that both PSNR curves increase as phase number $N_p$ increases; however, the curves are almost flat when $N_p\geq9$. Thus, considering the tradeoff between network complexity and reconstruction performance, we set the default phase number $N_p$=$9$ for both ISTA-Net and ISTA-Net$^+$ in the rest of the experiments. Clearly, ISTA-Net$^+$ achieves about 1 dB gain over ISTA-Net when $N_p$=$9$. Furthermore, Figure~\ref{fig: ISTA-Net vs ISTA-Net+} (bottom) plots the average PSNR curves for Set11 with respect to different numbers of epochs during training, when the CS ratio is 25\% and $N_p$=$9$. Both ISTA-Nets get higher PSNR when trained for a larger number of epochs, but ISTA-Net$^+$ achieves faster training convergence and better reconstruction performance on the test set (Set11). Due to limited space, please refer to \textbf{supplementary material} for the filters that are learned by ISTA-Nets.

We attribute the superiority of ISTA-Net$^+$ over ISTA-Net to two factors. First, ISTA-Net$^+$ explicitly sparsifies the images in the residual domain, leading to  a sparser representation as compared to ISTA-Net. Second, the skip connections introduced by ISTA-Net$^+$ coincide with the central idea of the popular ResNet \cite{he2016deep} architecture, which facilitates the training of deeper networks.

\subsection{Comparison with State-of-the-Art Methods}
We compare our proposed ISTA-Net and ISTA-Net$^+$ with five recent and state-of-the-art image CS methods, namely TVAL3 \cite{li2013efficient}, D-AMP \cite{metzler2016denoising}, IRCNN \cite{zhang2017learning}, SDA \cite{mousavi2015deep}, and ReconNet \cite{kulkarni2016reconnet}\footnote{We use the source code made  publicly  available  by the authors of TVAL3 \cite{li2013efficient}, D-AMP \cite{metzler2016denoising}, IRCNN \cite{zhang2017learning}, and ReconNet \cite{kulkarni2016reconnet} and implement SDA \cite{mousavi2015deep} ourselves, since its source code is not available.}. The first three are  optimization-based methods, while the last two are network-based methods. The average PSNR reconstruction performance on Set11 with respect to seven CS ratios are summarized in Table \ref{table: all PSNR Set11}. For fair comparison and following the evaluation strategy of  \cite{kulkarni2016reconnet}, all the competing methods reconstruct each image block from its CS measurement independently. From Table \ref{table: all PSNR Set11}, we observe that SDA  and ReconNet work better at extremely low CS ratios of 1\% and 4\%, while traditional optimization-based methods perform better at higher CS ratios. However, ISTA-Net and ISTA-Net$^+$ outperform all the existing methods by a large margin across all the CS ratios. This clearly  demonstrates that they combine the merits of both categories of CS methods. As expected, ISTA-Net$^+$ performs better than ISTA-Net. The last two columns in Table \ref{table: all PSNR Set11} is a run-time analysis of all the competing methods. These results indicate that the proposed ISTA-Nets produce consistently better reconstruction results, while remaining computationally attractive. In Figure~\ref{fig: visual}, we show the reconstructions of all seven methods of the \textit{Butterfly} image when the  CS ratio is 25\%. The proposed ISTA-Net$^+$ is able to reconstruct more details and sharper edges.

To further validate the generalizability of our ISTA-Nets, we also compare them to network-based methods SDA and ReconNet on the larger BSD68 dataset. As shown in Table \ref{table: all PSNR BSD68}, ISTA-Net$^+$ achieves the best performance, while ISTA-Net registers second best among all five CS ratios. ISTA-Nets outperform SDA and ReconNet, especially at higher CS ratios. In addition, it is worth emphasizing that a pre-trained ISTA-Net or ISTA-Net$^+$ using one CS measurement matrix $\mathbf{\Phi}$ can be directly used for any new measurement matrix with the same CS ratio as $\mathbf{\Phi}$, avoiding training new network from scratch. The only difference is that we need to recalculate the initialization matrix $\mathbf{Q}_{init}$ for the new measurement matrix, which usually takes less than 1 second. Please refer to \textbf{supplementary material} for more results.

\begin{table}[t]
\caption{Average PSNR (dB) performance comparison of various network-based algorithms on the BSD68 dataset.}
\centering
\small
\begin{tabular}{|l|c|c|c|c|c|}
\hline
\multirow{2}{*}{\textbf{Algorithm}} & \multicolumn{5}{c|}{\textbf{CS Ratio}}          \\ \cline{2-6}
                                                 & 50\%  & 40\%  & 30\%  & 10\%  & 4\%   \\ \hline \hline
SDA \cite{mousavi2015deep}                       & 28.35 & 27.41 & 26.38 & 23.12 & 21.32 \\
ReconNet \cite{kulkarni2016reconnet}             & 29.86 & 29.08 & 27.53 & 24.15 & 21.66 \\
ISTA-Net                                         & \underline{33.60} & \underline{31.85} & \underline{29.93} & \underline{25.02} & \underline{22.12} \\
ISTA-Net$^+$                                     & \textbf{34.01} & \textbf{32.21} & \textbf{30.34} & \textbf{25.33} & \textbf{22.17} \\ \hline
\end{tabular}
\label{table: all PSNR BSD68}
\vspace{-16pt}
\end{table}

\subsection{Comparison with ADMM-Net for CS-MRI}
\label{subsection:ADMM}
To demonstrate the generality of ISTA-Net$^+$, we directly extend  ISTA-Net$^+$ to the specific problem of CS MRI reconstruction, which aims at reconstructing MR images from a small number of under-sampled data in \textit{k}-space. In this application and following common practices, we set the sampling matrix $\mathbf{\Phi} $ in Eq.~(\ref{eq: classical CS}) to $\mathbf{\Phi} = \mathbf{PF}$, where $\mathbf{P}$ is an under-sampling matrix and $\mathbf{F}$ is the discrete Fourier transform. 
In this case, we compare against ADMM-Net \cite{yang2016deep}\footnote{https://github.com/yangyan92/Deep-ADMM-Net}, which is a network-based method inspired by ADMM and specifically designed for the CS-MRI domain. It is worthwhile to note that ADMM-Net cannot be trivially extended to other CS domains, since it imposes a specific  structure to the sampling matrix $\mathbf{\Phi}$. Utilizing the same training and testing brain medical images as ADMM-Net, the CS-MRI results of ISTA-Nets with $N_p$=11 phases are summarized in Table~\ref{table: ADMM-Net} for CS ratios of 20\%, 30\%, 40\% and 50\%. It is clear that ISTA-Nets outperform ADMM-Net not only in terms of reconstruction but also in terms of runtime.

\begin{figure}[t]
\centering
\includegraphics[width=0.4\textwidth]{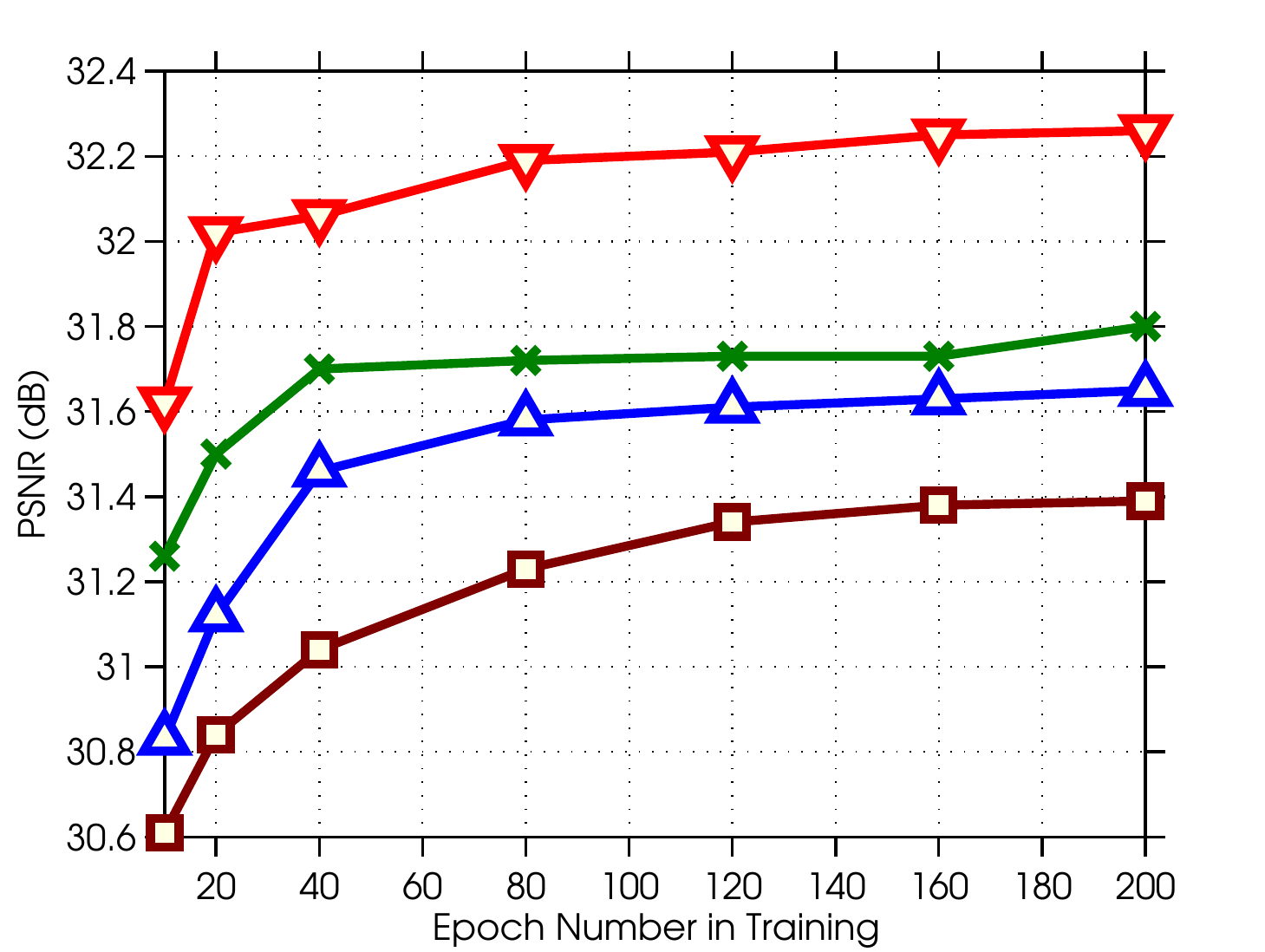}
\caption{PSNR (dB) comparison between two versions of ISTA-Net$^+$: with and without ReLU (when  $N_f$=$8$ and $N_f$=$16$).}
\label{fig: linear vs nonlinear}
\vspace{-4pt}
\end{figure}

\begin{table}[t]
\centering
\caption{Average PSNR (dB) comparison between ADMM-Net \cite{yang2016deep} and our proposed ISTA-Nets for CS-MRI.}
\small
\begin{tabular}{|l|c|c|c|c|c|}
\hline
\multirow{2}{*}{\textbf{Algorithm}} & \multicolumn{4}{c|}{\textbf{CS Ratio}} & \textbf{Time}   \\ \cline{2-5}
                           & 20\%  & 30\%  & 40\%  & 50\%  & GPU    \\ \hline \hline
ADMM-Net                   & 37.17 & 39.84 & 41.56 & 43.00 & 0.9535s \\
ISTA-Net                   & \underline{38.30} & \underline{40.52} & \underline{42.12} & \underline{43.60} & 0.1246s \\
ISTA-Net$^+$               & \textbf{38.73} & \textbf{40.89} & \textbf{42.52} & \textbf{44.09} & 0.1437s \\ \hline
\end{tabular}
\label{table: ADMM-Net}
\vspace{-4pt}
\end{table}

\begin{table}[t]
\centering
\caption{Average PSNR (dB) performance with different shared types of ISTA-Net$^+$.}
\small
\begin{tabular}{|l|r|c|}
\hline
\textbf{Shared Type}     & \textbf{Number of Parameters} & \textbf{PSNR} \\ \hline \hline
Shared $\rho^{(k)},\theta^{(k)},\mathcal{T}^{(k)}$ & (37440+1+1)=\textbf{37,442}    & 31.53     \\
Shared $\rho^{(k)},\mathcal{T}^{(k)}$    & (37440+1)+1*9=\underline{37,450}  & 32.28     \\
Shared $\theta^{(k)},\mathcal{T}^{(k)}$ & (37440+1)+1*9=\underline{37,450}  & 32.08     \\
Shared $\mathcal{T}^{(k)}$ & 37440+(1+1)*9=37,458  & \underline{32.36}     \\ \hline \hline
Unshared (default)       & (37440+1+1)*9=336,978 & \textbf{32.57}     \\ \hline
\end{tabular}
\label{table: shared type}
\vspace{-16pt}
\end{table}

\subsection{Ablation Studies and Discussions}
This section mainly focuses on the nonlinearity and flexibility of the proposed ISTA-Nets. In what follows, we analyze ISTA-Net$^+$ with $N_p$=$9$ phases.

\vspace{2pt}\noindent {$\bullet$~~ {\textbf{Linear~vs.~Nonlinear Transforms:~~}}}
The nonlinearity of ISTA-Net$^+$ is introduced by the ReLU operator in $\mathcal{H}^{(k)}$ and $\widetilde{\mathcal{H}}^{(k)}$, as shown in Figure~\ref{fig: ista-net-plus_phase}. To evaluate the impact of the nonlinearity, we train ISTA-Net$^+$ models with ReLU (nonlinear transforms) and without ReLU (linear transforms). Figure~\ref{fig: linear vs nonlinear} plots the average PSNR curves for each ISTA-Net$^+$ variant on Set11 throughout training. 
Note that parameter $N_f$, the number of feature maps in $\mathcal{H}^{(k)}$ and $\widetilde{\mathcal{H}}^{(k)}$, is set to $8$ or $16$ in this experiment. It is clear that the nonlinearity introduced by the ReLU is critical for high fidelity CS reconstruction performance. In addition, when $N_f>30$, experiments indicate that ISTA-Net$^+$ without ReLU is significantly less stable in training than ISTA-Net$^+$ with ReLU, which still performs well. We conclude that the nonlinearity plays an important role in facilitating satisfaction of the symmetry constraint, improving network stability, and learning a suitable transform possible for CS.

\vspace{2pt}
\noindent {$\bullet$~~ {\textbf{Shared~vs.~Unshared:~~}}} As described previously, each phase of ISTA-Net$^+$ ($N_f$=$32$) has three types of parameters with their dimensionality listed in parentheses: step size $\rho^{(k)}$ ({\small$1$}), threshold  $\theta^{(k)}$ ({\small $1$}), and transform $\mathcal{T}^{(k)} = \{\mathcal{D}^{(k)}, \mathcal{G}^{(k)}, \mathcal{H}^{(k)}, \widetilde{\mathcal{H}}^{(k)} \}$ {\small ($32\times3\times3+32\times3\times3\times32\times2+32\times3\times3\times32\times2+1\times3\times3\times32=37440$}). The flexibility of ISTA-Net$^+$ indicates that the same type of parameters in different phases do not need to be the same. To demonstrate the impact of this flexibility, we train several variants of ISTA-Net$^+$, where we vary the parameters that are shared among the phases. A summary of the average PSNR results on Set11 at a 25\% CS ratio  is reported in Table~\ref{table: shared type}. 
Obviously, the default \emph{unshared} ISTA-Net$^+$ (most flexible with largest number of parameters) achieves the best performance, while the variant of ISTA-Net$^+$ that shares all  parameters $(\rho^{(k)},\theta^{(k)},\mathcal{T}^{(k)})$ in all its phases (least flexible with  smallest number of parameters) obtains the worst performance. When only $(\rho^{(k)}, \mathcal{T}^{(k)})$ or $(\theta^{(k)}, \mathcal{T}^{(k)})$ are shared, these ISTA-Net$^+$  variants register 0.75dB and 0.55dB gains over  he variant with all shared parameters. Interestingly, the ISTA-Net$^+$ variant with only shared transforms $\mathcal{T}^{(k)}$ obtains very competitive PSNR results compared to the unshared variant. This indicates that further compression in ISTA-Net$^+$ parameters is possible, with limited affect on reconstruction performance.

\section{Conclusion and Future Work}
Inspired by the Iterative Shrinkage-Thresholding Algorithm (ISTA), we propose a novel structured deep network for image compressive sensing (CS) reconstruction, dubbed ISTA-Net, as well as, its enhanced version ISTA-Net$^+$. The proposed ISTA-Nets have well-defined interpretability, and make full use of the merits of both optimization-based and network-based CS methods. All the parameters in ISTA-Nets are discriminately learned end-to-end. Extensive experiments show that ISTA-Nets greatly improve upon the results of state-of-the-art CS methods, while  maintaining a fast runtime. Since the developed strategy to solve the proximal mapping problem associated to a nonlinear sparsifying transform is quite general and efficient, one direction of interest is to design deep networks based on other optimization inspirations, such as FISTA \cite{beck2009fast}. The other direction of our future work is to extend ISTA-Nets for other image inverse problems, such as deconvolution and inpainting.

\noindent \textbf{Acknowledgments.} This work was supported by the King Abdullah University of Science and Technology (KAUST) Office of Sponsored Research. The first author would like to sincerely thank Adel Bibi for his helpful discussion.

\end{document}